\documentclass[11pt]{article}

\usepackage[final]{acl}

\usepackage{times}
\usepackage{latexsym}

\usepackage[T1]{fontenc}

\usepackage[utf8]{inputenc}

\usepackage{microtype}

\usepackage{inconsolata}

\usepackage{graphicx}

\usepackage{booktabs}
\usepackage{multirow}
\usepackage{amsmath}
\usepackage{todonotes}

\title{Disentangling Linguistic Relatedness from Task Alignment in Cross-Lingual Transfer}

\author{
 \textbf{Ahmed Haj Ahmed\textsuperscript{1}},
 \textbf{Ruochen Zhang\textsuperscript{2}},
 \textbf{Alvin Grissom II\textsuperscript{1}}
\\
 \textsuperscript{1}Haverford College,
 \textsuperscript{2}Brown University
\\
 \small{
   \textbf{Correspondence:} \href{mailto:ahajahmed@haverford.edu}{ahajahmed@haverford.edu}
 }
}

\begin{document}
\maketitle
\begin{abstract}

We study cross-lingual transfer by fine-tuning seven large language models (4B--671B parameters) on Arabic and evaluating zero-shot reading comprehension on Semitic languages and non-Semitic controls. Across dense and Mixture-of-Experts architectures, we find no evidence of Semitic-specific transfer: models with weak baselines improve dramatically across all languages, while strong-baseline models show only marginal gains regardless of language family. A chain-of-thought ablation reinforces this finding: the same models that benefit most from fine-tuning benefit equally from inference-time reasoning, suggesting both mechanisms address task-format alignment rather than cross-lingual knowledge transfer.

\end{abstract}

\section{Introduction}

Large language models (LLMs) have dramatically advanced natural language processing, yet their benefits remain disproportionately concentrated in English and a small number of high-resource languages \citep{benderetal2021, joshi2021statefatelinguisticdiversity,bender2024power}. While recent frontier models demonstrate impressive multilingual coverage \citep{openai2025gptoss120bgptoss20bmodel, yang2025qwen3technicalreport, deepseekai2025deepseekv3technicalreport}, it remains unclear \emph{when and why} fine-tuning on one language yields performance gains, or failures, in another. Cross-lingual transfer, whereby capabilities learned by a model for one or more languages also benefit the model's capabilities in other languages, is often implicitly attributed to model scale or instruction tuning \citep{conneau-etal-2020-unsupervised, pires2019multilingualmultilingualbert}, yet growing evidence suggests that such factors alone are insufficient to explain transfer behavior across languages \citep{lauscher2020zeroherolimitationszeroshot, bankula2025crosslinguistictransfermultilingualnlp}. 

A natural testbed for studying cross-lingual transfer is the Semitic language family. Arabic, Hebrew, Amharic, and Maltese share morphological and syntactic properties~\cite{faber1997genetic}, including root-based non-concatenative morphology and rich inflection \citep{McCarthy1981APT, phonology_and_morphology_arabic}, while diverging substantially in script: Arabic and Hebrew use abjad systems, Amharic uses an abugida (Ge'ez), and Maltese uses the Latin alphabet. This combination of linguistic relatedness and orthographic diversity makes it possible to disentangle whether transfer operates at the level of linguistic structure, script similarity, or neither.

In this work, we fine-tune seven multilingual LLMs on a diverse set of Arabic dialects and evaluate zero-shot reading comprehension on Hebrew, Amharic, and Maltese, using Japanese, Korean, and French as non-Semitic controls. We use the Belebele benchmark \citep{Bandarkar_2024}, in which each item presents a short passage followed by a four-option multiple-choice question whose correct answer requires reasoning over information in the passage rather than surface lexical matching. Belebele provides parallel questions across all target languages, enabling strict control over train-test overlap. Our models span 4B to 671B parameters and cover both dense and Mixture-of-Experts (MoE) architectures, allowing us to examine how scale and architectural paradigm influence transfer patterns.

Our central finding is negative: we observe no evidence of Semitic-specific cross-lingual transfer. Models with weak zero-shot baselines---particularly the MoE models GPT-OSS-120B and GPT-OSS-20B---improve dramatically after Arabic fine-tuning, but gains are distributed relatively uniformly across all target languages, including non-Semitic controls. Models with strong baselines show only marginal improvement, regardless of language family. A chain-of-thought (CoT) reasoning ablation further supports this interpretation: the models that benefit most from fine-tuning are the same ones that benefit most from inference-time reasoning, and by comparable magnitudes. Together, these results suggest that Arabic fine-tuning primarily teaches task-format alignment rather than transferring language-specific knowledge.

Our contributions are as follows:
\begin{enumerate}
    \item We present a controlled cross-lingual transfer study across the Semitic language family, jointly varying linguistic relatedness, script, model scale, and architecture (dense vs.\ MoE), evaluated on parallel data.
    \item We present evidence that observed gains are not driven by language-family transfer, supported by regression analysis controlling for baseline accuracy.
    \item We introduce a CoT reasoning ablation, supported by a distributional calibration analysis (Appendix~\ref{sec:calibration_details}), showing that fine-tuning and inference-time reasoning compensate for the same underlying deficit in task-format calibration.
\end{enumerate}

\section{Related Work}

Our work lies at the intersection of cross-lingual transfer learning, Semitic language modeling, and architectural scaling in MoE models.

\subsection{Cross-Lingual Transfer and the ``Curse of Multilinguality''}

Cross-lingual transfer—the ability of a model to apply knowledge learned in a source language to a target language—has been a central objective of multilingual NLP, driven by models such as mBERT \citep{devlin2019bertpretrainingdeepbidirectional} and XLM-R \citep{conneau-etal-2020-unsupervised}. These encoder-based models demonstrate strong zero-shot transfer by inducing shared multilingual representations. However, the transition to decoder-only generative LLMs (e.g., XGLM, BLOOM) has introduced new challenges related to capacity allocation and interference \citep{lin2022fewshotlearningmultilinguallanguage, workshop2023bloom176bparameteropenaccessmultilingual}. Unlike encoder models trained primarily for representation alignment, decoder-only models must allocate the same parameters to next-token prediction across all languages, which can amplify competition between languages with different scripts, morphologies, and resource levels \citep{wang_etal_2020_negative, chang2023multilingualitycurselanguagemodeling}.

A recurring concern in multilingual modeling is the ``curse of multilinguality'' \citep{chang2023multilingualitycurselanguagemodeling}, whereby adding languages to a model of fixed capacity can dilute performance due to parameter competition. While increasing model scale can mitigate this effect, prior work has shown that fine-tuning on a single language or task may catastrophically overwrite multilingual representations in dense models \citep{phang2020englishintermediatetasktrainingimproves, lauscher2020zeroherolimitationszeroshot}. Our results extend this line of inquiry to MoE architectures, evaluating whether sparse expert routing mitigates such interference.

\subsection{Semitic Language Processing}

The Semitic language family presents persistent challenges for language models due to non-concatenative morphology, rich inflection, and substantial script variation, including abjad (Arabic, Hebrew), abugida (Amharic), and Latin (Maltese). Prior work has largely focused on encoder-based models and monolingual or bilingual settings, such as AraBERT for Arabic \citep{antoun2021araberttransformerbasedmodelarabic} and related adaptations.

Relatively little work has examined generative cross-lingual transfer within the Semitic family, particularly for reasoning-intensive tasks in decoder-only models. While grouping languages by family has been shown to improve transfer in translation and representation learning---e.g., NLLB \citep{nllbteam2022languageleftbehindscaling}---it remains unclear whether such benefits extend to zero-shot reasoning tasks. Moreover, languages such as Amharic and Maltese are often excluded from multilingual evaluations due to limited resources or script divergence. Our work explicitly includes Maltese as a controlled ablation, enabling an examination of syntax-level transfer independent of script overlap, an angle largely missing from prior Semitic NLP work.

\subsection{Instruction Tuning, Reasoning, and Task-Format Alignment}

Instruction tuning and reinforcement learning from human feedback (RLHF) have been shown to improve usability and alignment in LLMs \citep{bai2022traininghelpfulharmlessassistant}, but examining their effects on reasoning requires nuance. CoT can elicit latent reasoning abilities in sufficiently large models \citep{wei2023chainofthoughtpromptingelicitsreasoning}, yet alignment-focused fine-tuning may suppress intermediate reasoning signals, a phenomenon sometimes described as an ``alignment tax'' \citep{kirk2024understandingeffectsrlhfllm, lin2024mitigatingalignmenttaxrlhf}. Several recent methods aim to mitigate this tradeoff, either by reshaping the instruction-tuning objective \citep{fu2024dispersethenmergepushinglimitsinstruction, ren2024learningselfaligningrethinkinginstruction} or by isolating world knowledge from alignment updates via mixture-of-experts adaptation \citep{dou2024loramoealleviateworldknowledge}.

Our work connects these findings to cross-lingual transfer by testing whether fine-tuning and CoT reasoning provide complementary or redundant benefits. If both mechanisms address the same bottleneck—task-format alignment rather than linguistic knowledge—then the models that benefit from one should benefit equally from the other. Our reasoning ablation supports this prediction, suggesting that what appears to be cross-lingual transfer may in many cases reflect improved alignment with the evaluation format.

\section{Experimental Setup}

This section describes the data, models, fine-tuning procedure, and evaluation protocol used to study Arabic-to-target-language transfer under controlled conditions.

\subsection{Benchmark and Task}

We evaluate cross-lingual transfer using the Belebele benchmark \citep{Bandarkar_2024}, a massively parallel multiple-choice reading comprehension dataset spanning over 120 languages. Each example consists of a short passage, a question, and four candidate answers, designed to assess reasoning and comprehension rather than lexical overlap, following prior multilingual reading comprehension benchmarks \citep{joshi2017triviaqalargescaledistantly, zhang2018recordbridginggaphuman, clark2019boolqexploringsurprisingdifficulty, khashabi2020unifiedqacrossingformatboundaries}. Belebele is particularly well-suited for cross-lingual transfer studies because all languages share identical question identifiers, enabling strict control over train–test overlap across languages.

\subsection{Languages and Data Splits}

We fine-tune exclusively on Arabic dialects, using a total of 4,800 training examples drawn from six dialects: Mesopotamian, Tunisian, South Levantine, North Levantine, Najdi, and Moroccan Arabic. For each dialect, we sample 800 examples of the available 900 examples, reserving the remaining 100 per dialect for evaluation. Combining multiple dialects follows established practice in Arabic NLP \citep{bouamor-etal-2019-madar, abdul-mageed-etal-2021-arbert}, and exposes the model to a broader range of morphological and lexical variation than any single dialect would provide, reducing the risk of overfitting to dialect-specific surface patterns. All training examples are selected using a strict question ID–based split, ensuring that no parallel instances of the same question appear in any evaluation language.

We perform zero-shot evaluation on three Semitic target languages (Hebrew, Amharic, and Maltese) and three non-Semitic controls (Japanese, Korean, and French). This design follows established practice in cross-lingual transfer research, where unrelated languages are used to disentangle the effects of language-family transfer from those of general task adaptation \citep{singh-etal-2020-effects, Eronen_2023}. All evaluation sets are drawn from the Belebele validation split and contain no overlap with training data.

\subsection{Models}

We evaluate seven pretrained multilingual LLMs spanning a wide range of scales (4B--671B) and two architectural paradigms. Dense models activate all parameters for every input token, while Mixture-of-Experts (MoE) models contain many specialized parameter blocks (``experts'') and route each token to a small subset of them, so that only a fraction of the total parameter count is active per forward pass.

\begin{itemize}
    \item \textbf{Dense models}: Qwen3-4B-Instruct, Qwen3-8B, Qwen3-8B-Base, and Qwen3-32B \citep{yang2025qwen3technicalreport}.
    \item \textbf{MoE models}: GPT-OSS-20B, GPT-OSS-120B, and DeepSeek-V3.1 (671B) \citep{openai2025gptoss120bgptoss20bmodel, deepseekai2025deepseekv3technicalreport}.
\end{itemize}

The Qwen3 family enables controlled comparison across scale and training stage. Qwen3-8B-Base is a pretrained-only model with no post-training alignment, while Qwen3-8B and Qwen3-32B are post-trained models that support CoT reasoning \citep{wei2023chainofthoughtpromptingelicitsreasoning, nye2021workscratchpadsintermediatecomputation}.

MoE models are included to study how sparse expert routing---the mechanism by which a learned gating network sends each token to only a few of the model's experts---affects multilingual fine-tuning dynamics, motivated by evidence that expert specialization can mitigate interference across tasks and languages \citep{fedus2022switchtransformersscalingtrillion}.

\subsection{Fine-Tuning Procedure}
\label{finetuning}

All models are fine-tuned using supervised learning on the Arabic dialect training set, with infrastructure and distributed training managed via the \textsc{Tinker} training API \citep{tml2025tinker}.

\paragraph{Training format.}

Each training example is formatted as a single-turn conversation: the user turn contains the full Arabic prompt (passage, question, and four answer choices; see Appendix~\ref{sec:prompt_template} for the exact template), and the assistant turn contains the correct answer as a single digit (1--4). The prompt used during training is identical to the one used at evaluation time, ensuring no mismatch between training and test formats. Each model's recommended chat template is applied to wrap the conversation (e.g., \texttt{qwen3} for the Qwen3 family). The training loss is computed only over assistant-turn tokens, so the model is supervised exclusively on answer prediction. No CoT annotations or intermediate reasoning supervision are provided during fine-tuning.

\paragraph{Hyperparameters.}

We apply a uniform configuration across all models: LoRA with rank 32 \citep{hu2021loralowrankadaptationlarge}, three epochs, batch size 64, learning rate $2 \times 10^{-5}$ with linear decay, and Adam optimization \citep{kingma2017adammethodstochasticoptimization}. We use the final checkpoint (end of epoch 3) for all evaluations. Full hyperparameter details are provided in Appendix~\ref{sec:hyperparameters}.

No language-adaptive pretraining, multilingual mixing, or task-specific curriculum learning is applied during fine-tuning. This design isolates the effect of supervised adaptation on Arabic dialect data while minimizing confounding factors related to training procedure or optimization.

\subsection{Evaluation Protocol}
\label{evaluation}

We evaluate all models on the Belebele multiple-choice benchmark using accuracy. For each example, the prompt begins with a fixed instruction written in Arabic, followed by the passage, question, and four answer choices. While the instruction is always in Arabic, the passage, question, and answer choices are written in the evaluated target language (e.g., Hebrew, Amharic, Maltese, Japanese, Korean, or French). This design ensures that performance differences reflect cross-lingual transfer of comprehension skills rather than the ability to interpret instructions in different languages. The full prompt template is provided in Appendix~\ref{sec:prompt_template}.

\paragraph{Multiple-choice scoring.}

We score answers using probability rather than free-form generation. For each example, the prompt $c$ is passed through the model's chat template and tokenized. An anchor string $a = \texttt{``ANSWER:~''}$ is appended. We then score each numeric token $d \in D = \{1, 2, 3, 4\}$ by its conditional log-probability under the model:

\begin{equation}
    \text{score}(d) = \log p\!\left(d \mid c \circ a\right),
    \label{eq:logprob}
\end{equation}

where $\circ$ denotes token-sequence concatenation. The predicted answer is $\arg\max_{d \in D} \text{score}(d)$.

Each digit (1--4) maps to a single token in all tokenizers used in this study, which we verified by encoding each digit string without special tokens and taking the first resultant token ID. This scoring method avoids the ambiguity of parsing free-form text responses and ensures that evaluation is deterministic and comparable across models.

\paragraph{Inference configuration.}

All evaluations use greedy decoding with temperature set to 0.0, ensuring deterministic and reproducible outputs. Each model is evaluated using its recommended chat template (e.g., \texttt{qwen3} for the Qwen3 family), matching the template used during fine-tuning. Baseline and fine-tuned models are evaluated under identical conditions---same prompt format, decoding parameters, and scoring method---so that accuracy differences isolate the effect of fine-tuning. We report both absolute accuracy and accuracy change in percentage points.

\section{Results}
\label{sec:results}

Table~\ref{tab:finetuning} presents zero-shot accuracy on each target language before and after Arabic fine-tuning, along with the change in percentage points. Because each model is evaluated on the same 100 questions per language before and after fine-tuning, we can compute paired uncertainty bands for each gain; we report these in Appendix~\ref{sec:ci_details} for readers who want a sense of how tightly the per-question structure constrains the estimates. We treat them as descriptive bookkeeping rather than as significance tests, and the claims that follow rest on raw effect sizes.\footnote{The hedging here is intentional and is emphatically not an endorsement of the significance testing framework or of confidence intervals as a tool for inference. CIs are routinely misinterpreted as statements about a parameter's probable range~\citep{hoekstra2014robust}, which is fallacious despite being extremely common. CIs, however, are easy to compute and do contain uncertainty information, even if exactly what that information means in practical settings is difficult to interpret~\citep{gelman2019confidence}.} We organize our analysis around three main observations.

\begin{table*}[t]
\centering
\small
\begin{tabular}{llccccccc}
\toprule
\textbf{Model} & & \textbf{Amharic} & \textbf{Hebrew} & \textbf{Maltese} & \textbf{Japanese} & \textbf{Korean} & \textbf{French} & \textbf{Avg.} \\
\midrule
Random baseline & & 25.0 & 25.0 & 25.0 & 25.0 & 25.0 & 25.0 & 25.0 \\
\midrule
\multirow{3}{*}{DeepSeek-V3.1 (671B)} 
  & Base & 75.0 & 91.0 & 87.0 & 91.0 & 92.0 & 93.0 & 88.2 \\
  & FT   & 81.0 & 94.0 & 89.0 & 93.0 & 94.0 & 92.0 & 90.5 \\
  & $\Delta$ & +6.0 & +3.0 & +2.0 & +2.0 & +2.0 & $-$1.0 & +2.3 \\
\midrule
\multirow{3}{*}{GPT-OSS-120B} 
  & Base & 30.0 & 41.0 & 32.0 & 55.0 & 56.0 & 58.0 & 45.3 \\
  & FT   & 52.0 & 85.0 & 73.0 & 84.0 & 88.0 & 91.0 & 78.8 \\
  & $\Delta$ & +22.0 & +44.0 & +41.0 & +29.0 & +32.0 & +33.0 & +33.5 \\
\midrule
\multirow{3}{*}{GPT-OSS-20B} 
  & Base & 27.0 & 66.0 & 29.0 & 69.0 & 78.0 & 82.0 & 58.5 \\
  & FT   & 50.0 & 91.0 & 73.0 & 89.0 & 91.0 & 89.0 & 80.5 \\
  & $\Delta$ & +23.0 & +25.0 & +44.0 & +20.0 & +13.0 & +7.0 & +22.0 \\
\midrule
\multirow{3}{*}{Qwen3-32B} 
  & Base & 52.0 & 87.0 & 85.0 & 93.0 & 90.0 & 91.0 & 83.0 \\
  & FT   & 55.0 & 88.0 & 86.0 & 94.0 & 90.0 & 91.0 & 84.0 \\
  & $\Delta$ & +3.0 & +1.0 & +1.0 & +1.0 & 0.0 & 0.0 & +1.0 \\
\midrule
\multirow{3}{*}{Qwen3-8B} 
  & Base & 44.0 & 79.0 & 65.0 & 82.0 & 89.0 & 92.0 & 75.2 \\
  & FT   & 44.0 & 79.0 & 63.0 & 84.0 & 89.0 & 92.0 & 75.2 \\
  & $\Delta$ & 0.0 & 0.0 & $-$2.0 & +2.0 & 0.0 & 0.0 & 0.0 \\
\midrule
\multirow{3}{*}{Qwen3-8B-Base} 
  & Base & 43.0 & 81.0 & 72.0 & 83.0 & 88.0 & 89.0 & 76.0 \\
  & FT   & 45.0 & 85.0 & 70.0 & 83.0 & 87.0 & 91.0 & 76.8 \\
  & $\Delta$ & +2.0 & +4.0 & $-$2.0 & 0.0 & $-$1.0 & +2.0 & +0.8 \\
\midrule
\multirow{3}{*}{Qwen3-4B-Instruct} 
  & Base & 37.0 & 81.0 & 54.0 & 84.0 & 90.0 & 91.0 & 72.8 \\
  & FT   & 42.0 & 81.0 & 61.0 & 82.0 & 89.0 & 90.0 & 74.2 \\
  & $\Delta$ & +5.0 & 0.0 & +7.0 & $-$2.0 & $-$1.0 & $-$1.0 & +1.3 \\
\bottomrule
\end{tabular}
\caption{Zero-shot accuracy (\%) on target languages before (Base) and after (FT) Arabic dialect fine-tuning. $\Delta$ denotes the change in percentage points. Semitic target languages are Amharic, Hebrew, and Maltese; non-Semitic controls are Japanese, Korean, and French.}
\label{tab:finetuning}
\end{table*}

\paragraph{MoE models with weak baselines benefit most.}

The most striking result is the scale of improvement in the two GPT-OSS models. GPT-OSS-120B improves by an average of +33.5 percentage points, rising from 45.3\% to 78.8\%, while GPT-OSS-20B gains +22.0 pp on average, rising from 58.5\% to 80.5\%. All individual languages show convincing gains. These gains are not concentrated on Semitic languages: for GPT-OSS-120B, French improves by +33.0 pp, Korean by +32.0 pp, and Japanese by +29.0 pp---of the same order as the gains on Hebrew (+44.0 pp), Maltese (+41.0 pp), and Amharic (+22.0 pp), with no consistent Semitic advantage. The pattern for GPT-OSS-20B is similar, with Maltese showing the largest single gain (+44.0 pp) but Japanese (+20.0 pp) and Korean (+13.0 pp) also improving substantially. The uniformity of these gains across language families strongly suggests that Arabic fine-tuning is teaching task-format alignment rather than transferring Semitic-specific linguistic knowledge. An analysis of the underlying logprob distributions confirms that these accuracy gains correspond to distributional sharpening toward correct answers, with magnitude tracking baseline entropy rather than language family (Appendix~\ref{sec:calibration_details}).

\paragraph{Strong baselines leave little room for improvement.}

At the other end of the spectrum, DeepSeek-V3.1 begins at 88.2\% average accuracy and improves by only +2.3 pp after fine-tuning. Qwen3-32B, starting at 83.0\%, gains just +1.0 pp. Most per-language gains for these models are within a few points and within the range of what paired uncertainty bands can distinguish from zero (Appendix~\ref{sec:ci_details}); the one clear exception is Amharic for DeepSeek-V3.1 (+6.0 pp). The takeaway is that these models already perform well on the Belebele task format, and fine-tuning yields only marginal improvement.

\paragraph{Dense 8B models show near-zero transfer.}

The three dense models at or near the 8B scale---Qwen3-8B, Qwen3-8B-Base, and Qwen3-4B-Instruct---show minimal response to Arabic fine-tuning, with average gains of 0.0, +0.8, and +1.3 pp respectively, changes small enough to be indistinguishable from noise on a 100-item evaluation. Qwen3-8B is particularly notable: accuracy is completely unchanged on four of six languages, with a $-$2.0 pp drop on Maltese offset by a +2.0 pp gain on Japanese. Qwen3-4B-Instruct shows a mixed pattern, with modest gains on Amharic (+5.0 pp) and Maltese (+7.0 pp) but slight degradation on Japanese ($-$2.0 pp), Korean ($-$1.0 pp), and French ($-$1.0 pp). This suggests that at small dense model scales, LoRA fine-tuning on Arabic data provides insufficient signal to move the needle on cross-lingual reading comprehension, and may in some cases introduce mild catastrophic forgetting on high-resource languages.

\paragraph{Regression analysis finds no Semitic-specific effect.}

To test whether language-family membership provides residual explanatory power beyond baseline accuracy, we fit an ordinary least squares linear regression of fine-tuning gain (in percentage points) on baseline accuracy and a binary Semitic indicator across all 42 model--language observations (full details in Appendix~\ref{sec:regression}). Baseline accuracy alone explains nearly half of the variance in fine-tuning gains ($R^2=0.479$, $\beta_{\text{baseline}}=-0.45$, $t=-6.07$). Adding the Semitic indicator yields $\Delta R^2=0.065$, but the Semitic coefficient is \emph{negative}: $\beta_{\text{Semitic}}=-8.2$ pp ($t=-2.35$, $\text{SE}=3.50$). That is, after controlling for baseline accuracy, Semitic languages gain \emph{less} from Arabic fine-tuning than non-Semitic controls---the opposite of what language-family transfer would predict. This reverses because language family is confounded with baseline accuracy: Semitic targets have substantially lower mean baselines (59.9\% vs.\ 82.7\%), inflating their raw gains through greater headroom. These results provide no evidence that Arabic fine-tuning preferentially benefits typologically related languages.

\paragraph{Script similarity does not predict transfer.}

A related hypothesis is that transfer operates at the level of script rather than language family: since Arabic fine-tuning exposes the model to abjad-script input, Hebrew (also an abjad) might benefit disproportionately relative to languages written in abugida (Amharic), Latin (Maltese, French), or CJK scripts (Japanese, Korean). To test this, we extend the regression analysis in two ways (full details in Appendix~\ref{sec:regression}). First, we replace the Semitic indicator with a binary abjad indicator (1 for Hebrew, 0 otherwise). The abjad coefficient is small ($\beta_{\text{abjad}} = +4.85$ pp, $t = 1.17$, $\text{SE} = 4.16$; model $R^2 = 0.497$), providing no evidence that sharing a script type with Arabic confers a transfer advantage. Second, we code an ordinal script-distance variable (abjad = 0, abugida = 1, Latin = 2, CJK = 3), capturing a gradient of orthographic similarity to Arabic. This predictor is likewise small ($\beta_{\text{dist}} = +1.56$ pp per step, $t = 1.02$, $\text{SE} = 1.52$; model $R^2 = 0.493$). Neither script-level predictor improves meaningfully over the baseline-only model ($R^2 = 0.479$), indicating that orthographic overlap with the fine-tuning language does not drive transfer. Per-model inspection reinforces this conclusion: headroom-normalized gains for the abjad group (Hebrew) are 0.75 and 0.74 for GPT-OSS-120B and GPT-OSS-20B respectively, compared to 0.69 and 0.50 for Latin-script languages and 0.69 and 0.62 for CJK---differences that are small, inconsistent in direction across models, and fully accounted for by baseline variation.

\section{Reasoning Ablation}
\label{sec:reasoning_ablation}

To further probe the nature of the gains observed in Section~\ref{sec:results}, we conduct an ablation comparing two scoring regimes on the \emph{baseline} (non-fine-tuned) models. The standard regime computes $\text{score}(d)$ from Eq.~\ref{eq:logprob}. The reasoning regime first generates a free-form CoT response $r$ using greedy decoding ($\tau=0$) with a maximum of 4,096 tokens, then scores:

\begin{equation}
    \text{score}_{\text{CoT}}(d) = \log p\!\left(d \mid c \circ r \circ a\right).
    \label{eq:logprob_cot}
\end{equation}

That is, the generated reasoning trace $r$ is inserted between the original prompt and the same anchor $a$ before scoring each digit, so the only difference from the standard regime is the additional CoT context. Truncation at the 4,096-token limit is rare (overall 3.6\% of examples; per-model rates in Appendix~\ref{sec:truncation}). Qwen3-4B-Instruct and Qwen3-8B-Base are omitted from this ablation as they do not support CoT inference.

If fine-tuning primarily injects cross-lingual knowledge, reasoning at inference time should not replicate its effects. If, on the other hand, both mechanisms address the same bottleneck---task-format alignment---then the models that benefit from fine-tuning should benefit equally from reasoning.
Table~\ref{tab:reasoning} presents the results. We highlight three findings.

\begin{table*}[t]
\centering
\small
\begin{tabular}{llccccccc}
\toprule
\textbf{Model} & & \textbf{Amharic} & \textbf{Hebrew} & \textbf{Maltese} & \textbf{Japanese} & \textbf{Korean} & \textbf{French} & \textbf{Avg.} \\
\midrule
Random baseline & & 25.0 & 25.0 & 25.0 & 25.0 & 25.0 & 25.0 & 25.0 \\
\midrule
\multirow{3}{*}{DeepSeek-V3.1 (671B)}
  & No CoT  & 75.0 & 91.0 & 87.0 & 91.0 & 92.0 & 93.0 & 88.2 \\
  & With CoT & 75.0 & 87.0 & 88.0 & 88.0 & 94.0 & 93.0 & 87.5 \\
  & $\Delta$ & 0.0 & $-$4.0 & +1.0 & $-$3.0 & +2.0 & 0.0 & $-$0.7 \\
\midrule
\multirow{3}{*}{GPT-OSS-120B}
  & No CoT  & 30.0 & 41.0 & 32.0 & 55.0 & 56.0 & 58.0 & 45.3 \\
  & With CoT & 61.0 & 88.0 & 90.0 & 92.0 & 91.0 & 96.0 & 86.3 \\
  & $\Delta$ & +31.0 & +47.0 & +58.0 & +37.0 & +35.0 & +38.0 & +41.0 \\
\midrule
\multirow{3}{*}{GPT-OSS-20B}
  & No CoT  & 27.0 & 66.0 & 29.0 & 69.0 & 78.0 & 82.0 & 58.5 \\
  & With CoT & 48.0 & 86.0 & 81.0 & 88.0 & 91.0 & 94.0 & 81.3 \\
  & $\Delta$ & +21.0 & +20.0 & +52.0 & +19.0 & +13.0 & +12.0 & +22.8 \\
\midrule
\multirow{3}{*}{Qwen3-32B}
  & No CoT  & 52.0 & 87.0 & 85.0 & 93.0 & 90.0 & 91.0 & 83.0 \\
  & With CoT & 49.0 & 86.0 & 90.0 & 89.0 & 93.0 & 93.0 & 83.3 \\
  & $\Delta$ & $-$3.0 & $-$1.0 & +5.0 & $-$4.0 & +3.0 & +2.0 & +0.3 \\
\midrule
\multirow{3}{*}{Qwen3-8B}
  & No CoT  & 44.0 & 79.0 & 65.0 & 82.0 & 89.0 & 92.0 & 75.2 \\
  & With CoT & 44.0 & 84.0 & 81.0 & 85.0 & 93.0 & 90.0 & 79.5 \\
  & $\Delta$ & 0.0 & +5.0 & +16.0 & +3.0 & +4.0 & $-$2.0 & +4.3 \\
\bottomrule
\end{tabular}
\caption{Accuracy (\%) on baseline (non-fine-tuned) models with and without CoT reasoning (max 4,096 thinking tokens). $\Delta$ denotes the change in percentage points from adding CoT. Qwen3-4B-Instruct and Qwen3-8B-Base are omitted as they do not support CoT inference.}
\label{tab:reasoning}
\end{table*}

\paragraph{Reasoning and fine-tuning address the same bottleneck.}

The most important result of this ablation is the close correspondence between reasoning gains and fine-tuning gains. GPT-OSS-120B improves by +41.0 pp with CoT reasoning, compared to +33.5 pp from fine-tuning; GPT-OSS-20B gains +22.8 pp from reasoning versus +22.0 pp from fine-tuning. In both cases, the gains are distributed broadly across all languages rather than concentrated on Semitic targets. This parallel strongly suggests that both mechanisms compensate for the same underlying deficit: these models possess multilingual reading comprehension knowledge but fail to express it through direct logprob scoring without either explicit reasoning or supervised format adaptation. Distributional calibration metrics corroborate this pattern (Appendix~\ref{sec:calibration_details}).

\paragraph{Strong models gain little or are slightly hurt by reasoning.}

DeepSeek-V3.1 drops $-$0.7 pp on average when CoT is prepended, with a $-$4.0 pp decrease on Hebrew and $-$3.0 pp on Japanese partially offset by a +2.0 pp gain on Korean. Qwen3-32B shows a similarly flat pattern at +0.3 pp, with gains on Maltese (+5.0 pp) and Korean (+3.0 pp) counterbalanced by drops on Japanese ($-$4.0 pp) and Amharic ($-$3.0 pp). For models that are already well-calibrated on the task, prepending a reasoning trace appears to introduce noise into the logprob distribution without systematically improving answer selection.

\paragraph{Mid-scale dense models benefit moderately.}

Qwen3-8B gains +4.3 pp on average from reasoning, with a notable +16.0 pp jump on Maltese---the language on which its accuracy is weakest (65.0\%). This is a case where fine-tuning yields no average improvement (0.0 pp) but reasoning does help, suggesting that Qwen3-8B possesses relevant knowledge that LoRA fine-tuning on Arabic fails to activate but that explicit reasoning can elicit.

\section{Discussion}

Our results present a clear but perhaps counterintuitive picture: fine-tuning on Arabic data does not induce measurable Semitic-specific cross-lingual transfer. Instead, the dominant mechanism is task-format alignment, and the magnitude of improvement is almost entirely predicted by how poorly a model performs at baseline. 

\paragraph{Task-format alignment, not linguistic transfer.}

The strongest evidence against language-family transfer is the uniformity of gains across Semitic and non-Semitic targets. After controlling for baseline accuracy, the Semitic coefficient is negative ($-$8.2 pp, $t = -2.35$), indicating that Semitic languages benefit \emph{less} than non-Semitic controls once headroom differences are accounted for. The reasoning ablation provides converging evidence: CoT produces a nearly identical pattern of gains across the same models and languages, yet involves no parameter updates and no Arabic-specific training signal. A distributional calibration analysis corroborates this---both interventions sharpen answer-choice distributions toward correct answers for poorly calibrated models while leaving already-peaked distributions unchanged (Appendix~\ref{sec:calibration_details}). The most parsimonious explanation is that both fine-tuning and reasoning help models express knowledge they already possess but cannot surface through direct logprob scoring. This connects to the ``alignment tax'' literature \citep{bai2022traininghelpfulharmlessassistant, lin2024mitigatingalignmenttaxrlhf}, but inverts its framing: \emph{insufficient} format alignment---not insufficient knowledge---is the primary bottleneck for weaker models.

\paragraph{Architecture matters more than language family.}

A second implication is that architectural paradigm and baseline capability are stronger predictors of fine-tuning benefit than any property of the target language. The MoE models (GPT-OSS-120B, GPT-OSS-20B) show the largest gains, but this appears to reflect their weak initial task-format calibration rather than any advantage of sparse expert routing for cross-lingual transfer. DeepSeek-V3.1, also an MoE model but with a much stronger baseline, shows only marginal improvement. Among dense models, scale alone does not predict benefit: Qwen3-32B (83.0\% baseline) gains +1.0 pp, while Qwen3-8B (75.2\% baseline) gains 0.0 pp, suggesting that once a model reaches a threshold of format alignment, additional Arabic fine-tuning via LoRA provides diminishing returns regardless of scale.

This result echoes findings on the ``curse of multilinguality'' \citep{chang2023multilingualitycurselanguagemodeling}, but with an important nuance. The concern in prior work is that fixed-capacity models lose per-language performance as language coverage increases. Our results suggest a different dynamic: for well-aligned models, the bottleneck is not capacity but rather the marginal irrelevance of additional format-level supervision. The calibration analysis underscores this for small dense models: Qwen3-4B-Instruct shows a mean entropy \emph{increase} of +0.42 bits after fine-tuning, with $\Delta P^* \approx 0$, indicating that LoRA on Arabic data actively diffuses the logprob distributions of already-calibrated small models without redirecting mass toward correct answers.

\paragraph{Script overlap is not a transfer mechanism.}

Our experimental design deliberately varies script across evaluation languages, enabling a test that most cross-lingual studies leave implicit. If Arabic fine-tuning transferred knowledge through shared orthographic representations, Hebrew---which uses the same abjad script type and right-to-left directionality as Arabic---should benefit disproportionately. It does not. After controlling for baseline accuracy, neither a binary abjad indicator nor an ordinal script-distance gradient---which codes each script by its rough orthographic distance from Arabic's abjad (abjad = 0, abugida = 1, Latin = 2, CJK = 3)---adds meaningful explanatory power (Section~\ref{sec:results}). This result is consistent with recent findings that subword tokenizers in large multilingual models largely abstract away script-level features \citep{conneau-etal-2020-unsupervised}, and suggests that script similarity should not be used as a proxy for expected transfer benefit when selecting source languages for fine-tuning.

\paragraph{Implications for low-resource language strategies.}

A practical motivation for studying cross-lingual transfer is the hope that fine-tuning on a related high-resource language can bootstrap performance on lower-resource targets. Our results temper this hope, at least for reading comprehension tasks evaluated via logprob scoring. The gains we observe on Amharic---the lowest-resource language in our evaluation---are real (+6.0 pp for DeepSeek-V3.1, +22.0 pp for GPT-OSS-120B), but they are not preferentially larger than gains on high-resource controls. This suggests that practitioners seeking to improve performance on low-resource Semitic languages through Arabic fine-tuning should not expect language-family relatedness to provide a special advantage. Instead, the more reliable lever appears to be improving general task-format alignment, whether through fine-tuning, reasoning, or both.

That said, our experiments are anchored to multiple-choice reading comprehension scored via logprob ranking—a paradigm widely used in multilingual evaluation. Semitic-specific transfer may well emerge in tasks where shared root morphology plays a more direct role, such as morphological inflection, lemmatization, or open-ended generation, and we view such evaluations as a natural next step. Our finding is therefore best read as a methodological caution: gains observed on multiple-choice multilingual benchmarks should not be uncritically attributed to cross-lingual knowledge transfer without controls for task-format alignment.

\section{Conclusion}

We study cross-lingual transfer by fine-tuning seven multilingual language models (4B--671B parameters) on Arabic dialect data and evaluating zero-shot reading comprehension on Semitic and non-Semitic target languages. Our central finding is that Arabic fine-tuning does not induce Semitic-specific transfer. Models with weak baselines improve dramatically, but gains are distributed uniformly across all target languages, including non-Semitic controls, indicating that the primary effect is task-format alignment rather than the activation of shared linguistic representations.

A CoT reasoning ablation provides converging evidence for this interpretation. The models that benefit most from fine-tuning are the same ones that benefit most from inference-time reasoning, and by comparable magnitudes, suggesting both mechanisms compensate for the same underlying deficit in task-format calibration. An analysis of answer-choice logprob distributions corroborates this: both interventions sharpen distributions toward correct answers for poorly calibrated models while leaving already-peaked distributions unchanged or mildly disrupted. For models that are already well-calibrated, neither intervention yields substantial improvement.

These results carry practical implications for multilingual NLP. Practitioners seeking to improve performance on low-resource languages through related-language fine-tuning should not assume that language-family relatedness confers a special advantage. More broadly, our findings suggest that distinguishing between genuine cross-lingual knowledge transfer and task-format alignment is essential for interpreting fine-tuning gains in multilingual settings.

\section*{Limitations}

Our study has several limitations that should be considered when interpreting the results.

First, we evaluate on a single task format: multiple-choice reading comprehension scored via logprob ranking. It is possible that Semitic-specific transfer manifests in other settings, such as open-ended generation, morphological analysis, or tasks where shared root structure plays a more direct role. Relatedly, our prompts use a fixed Arabic instruction at both training and evaluation time; learning to follow that instruction may itself contribute to the alignment gains we report, though we view this as consistent with rather than contradictory to our central claim. Our negative finding should therefore be understood as specific to this evaluation paradigm.

Second, we fine-tune exclusively on Arabic, chosen as the highest-resource Semitic language with a rich multi-dialect training set. The asymmetry of this design means our findings most directly speak to Arabic-as-source transfer, and complementary experiments using Hebrew or Amharic as source languages, or symmetric multi-source fine-tuning, would broaden the scope of conclusions.

Third, our evaluation sets are drawn from the Belebele benchmark and are limited in size (100 examples per language after the question-ID split). We mitigate this by reporting paired uncertainty bands in Appendix~\ref{sec:ci_details} that exploit per-question structure. Nonetheless, several per-language gains for the dense models are not distinguishable from zero by these bands.

Fourth, we use a single fine-tuning configuration (LoRA rank 32, three epochs, uniform hyperparameters) across all models, and our CoT ablation uses an unconstrained reasoning protocol whose generated traces frequently contain an explicit answer indicator (Appendix~\ref{sec:calibration_details}). We partially address the latter through $\Delta P^*$ analysis, which only inflates when reasoning lands on the correct answer, but a leakage-free CoT protocol and model-specific hyperparameter tuning would each strengthen the corresponding conclusions.

Finally, we do not analyze expert routing patterns in the MoE models, which could shed light on whether Arabic fine-tuning activates language-specific or task-general experts. We leave this analysis to future work.

\section*{Acknowledgments}

We thank Zheng-Xin Yong for valuable feedback on this work. We are grateful to Thinking Machines for providing \$150 in credits through the \textsc{Tinker} API, which supports the fine-tuning infrastructure used in our experiments. We also thank the authors of the Belebele benchmark \citep{Bandarkar_2024}, whose massively parallel multilingual reading comprehension dataset makes this cross-lingual transfer study possible. We gratefully acknowledge the Marian E.~Koshland Integrated Natural Sciences Center at Haverford College for their funding support. Grissom is funded by NSF grant 2403439. We are also deeply appreciative of the three anonymous reviewers, whose constructive feedback and suggestions helped us strengthen our work.

\bibliography{custom}

\appendix

\section{Hyperparameter Details}
\label{sec:hyperparameters}

All models are fine-tuned with a uniform configuration. Training uses a maximum input length of 2,048 tokens, batch size of 64, and a learning rate of $2 \times 10^{-5}$ with linear decay and no warmup. Optimization uses Adam \citep{kingma2017adammethodstochasticoptimization} with $\beta_1 = 0.9$, $\beta_2 = 0.95$, and $\epsilon = 10^{-8}$. We employ LoRA with rank 32 \citep{hu2021loralowrankadaptationlarge} for parameter-efficient fine-tuning. Validation loss is evaluated every 50 optimization steps on a held-out set of 600 Arabic examples (questions 801--900 across all six dialects), and checkpoints are saved every 100 steps.

\section{Regression Analysis Details}
\label{sec:regression}

This appendix reports the full regression analysis summarized in Section~\ref{sec:results}. All regressions are estimated by ordinary least squares over $N = 42$ model--language observations (7 models $\times$ 6 target languages). The dependent variable is the fine-tuning gain $\Delta$ (in percentage points), and the predictors are baseline accuracy and a binary Semitic indicator (1 for Amharic, Hebrew, Maltese; 0 for Japanese, Korean, French).

\paragraph{Model 1: Gain $\sim$ Baseline.}
A simple linear regression of gain on baseline accuracy yields:
\begin{align}
    \widehat{\Delta} &= 40.65 - 0.448 \times \text{Baseline} \notag \\
    R^2 &= 0.479, \quad \beta_{\text{baseline}}\; t = -6.07,\; \text{SE} = 0.074 \notag
\end{align}
Baseline accuracy alone explains 47.9\% of the variance in fine-tuning gains, confirming that the dominant predictor is how much room a model--language pair has to improve.

\paragraph{Model 2: Gain $\sim$ Baseline + Semitic.}
Adding a binary Semitic indicator yields:
\begin{align}
    \widehat{\Delta} &= 52.34 - 0.554 \times \text{Baseline} - 8.23 \times \text{Semitic} \notag \\
    R^2 &= 0.544, \quad \Delta R^2 = 0.065 \notag
\end{align}
The Semitic coefficient is negative ($\beta_{\text{Semitic}} = -8.23$, $t = -2.35$, $\text{SE} = 3.50$), indicating that after controlling for baseline accuracy, Semitic target languages gain \emph{less} from Arabic fine-tuning than non-Semitic controls. The incremental variance explained by the Semitic indicator is modest ($\Delta R^2 = 0.065$) and runs counter to the direction predicted by language-family transfer.

\paragraph{Partial correlation.}
The partial correlation between fine-tuning gain and Semitic membership, controlling for baseline accuracy, is $r = -0.353$ ($r^2 = 0.124$, $t = -2.35$, $df = 39$).

\paragraph{Model 3: Gain $\sim$ Baseline + Abjad.}
To test whether sharing a script type with Arabic (abjad) predicts transfer, we replace the Semitic indicator with a binary abjad indicator (1 for Hebrew, 0 for all other targets):
\begin{align}
    \widehat{\Delta} &= 40.35 - 0.455 \times \text{Baseline} + 4.85 \times \text{Abjad} \notag \\
    R^2 &= 0.497, \quad \beta_{\text{abjad}}\; t = 1.17,\; \text{SE} = 4.16 \notag
\end{align}
The abjad coefficient is positive but small, and the model's $R^2$ barely exceeds the baseline-only model (0.479). Sharing an abjad script with Arabic does not predict fine-tuning benefit.

\paragraph{Model 4: Gain $\sim$ Baseline + Script Distance.}
We also test an ordinal script-distance variable coded as: abjad (Hebrew) = 0, abugida (Amharic) = 1, Latin (Maltese, French) = 2, CJK (Japanese, Korean) = 3:

\begin{align}
    \widehat{\Delta} &= 39.47 - 0.471 \times \text{Baseline} \notag \\
        &\quad {}+ 1.56 \times \text{ScriptDist} \notag \\
    R^2 &= 0.493, \quad \beta_{\text{dist}}\; t = 1.02,\; \text{SE} = 1.52 \notag
\end{align}

The script-distance coefficient is small and has the \emph{opposite} sign of the script-proximity hypothesis (positive rather than negative), indicating that more distant scripts do not transfer less well once baseline is controlled. Neither script-level predictor adds meaningful explanatory power beyond baseline accuracy.

\paragraph{Raw vs.\ adjusted group means.}
Table~\ref{tab:group_means} reports unadjusted and baseline-adjusted mean gains. The raw Semitic advantage (+4.38 pp) reverses after adjustment ($-$8.23 pp), reflecting the substantially lower Semitic baselines (59.9\% vs.\ 82.7\%) that provide more headroom.

\begin{table}[h]
\centering
\small
\begin{tabular}{lcc}
\toprule
& \textbf{Semitic} & \textbf{Non-Semitic} \\
\midrule
Mean baseline (\%) & 59.9 & 82.7 \\
Mean gain (raw, pp) & +10.90 & +6.52 \\
Raw difference & \multicolumn{2}{c}{+4.38 pp (Semitic $-$ Non-Semitic)} \\
\midrule
Adjusted gain (pp)\textsuperscript{$\dagger$} & +4.60 & +12.83 \\
Adjusted difference & \multicolumn{2}{c}{$-$8.23 pp (Semitic $-$ Non-Semitic)} \\
\bottomrule
\end{tabular}
\caption{Raw and adjusted mean fine-tuning gains by language family. \textsuperscript{$\dagger$}Adjusted means are evaluated at the grand mean baseline (71.3\%) using the coefficients from Model~2.}
\label{tab:group_means}
\end{table}

\paragraph{Per-model breakdown.}
Table~\ref{tab:per_model_regression} reports Semitic vs.\ non-Semitic gains for each model, along with a headroom-normalized metric (gain divided by available headroom, i.e., $100 - \text{baseline}$). Across models, the normalized gain difference between Semitic and non-Semitic targets is inconsistent in sign and small in magnitude, further supporting the conclusion that language-family membership does not systematically predict fine-tuning benefit.

\begin{table}[h]
\centering
\small
\begin{tabular}{lcccc}
\toprule
\textbf{Model} & \multicolumn{2}{c}{\textbf{Avg.\ Gain (pp)}} & \multicolumn{2}{c}{\textbf{Gain / Headroom}} \\
\cmidrule(lr){2-3} \cmidrule(lr){4-5}
& Sem. & Non-S. & Sem. & Non-S. \\
\midrule
DeepSeek-V3.1  & +3.7  & +1.0  & 0.234 & 0.125 \\
GPT-OSS-120B   & +35.7 & +31.3 & 0.543 & 0.718 \\
GPT-OSS-20B    & +30.7 & +13.3 & 0.517 & 0.563 \\
Qwen3-32B      & +1.7  & +0.3  & 0.066 & 0.038 \\
Qwen3-4B-Inst. & +4.0  & $-$1.3 & 0.094 & $-$0.114 \\
Qwen3-8B       & $-$0.7 & +0.7 & $-$0.018 & 0.054 \\
Qwen3-8B-Base  & +1.3  & +0.3  & 0.038 & 0.025 \\
\bottomrule
\end{tabular}
\caption{Per-model fine-tuning gains for Semitic vs.\ non-Semitic targets. Gain/Headroom normalizes by available improvement room ($100 - \text{baseline}$). The normalized difference is inconsistent in sign across models, indicating no systematic Semitic advantage.}
\label{tab:per_model_regression}
\end{table}

\section{Confidence Interval Details}
\label{sec:ci_details}

This appendix reports paired 95\% confidence intervals for all fine-tuning gains using McNemar's test, a paired comparison for binary outcomes that exploits the fact that the same questions are answered before and after fine-tuning. Compared to unpaired binomial comparisons, this yields substantially tighter intervals (paired standard errors are 18--91\% of their unpaired counterparts, depending on the model). For each model--language pair evaluated on $n = 100$ shared questions, we construct a $2 \times 2$ contingency table of per-question correctness under the baseline and fine-tuned models. Let $b$ denote questions the baseline answered correctly but the fine-tuned model did not (regressions), and $c$ denote the reverse (improvements). The net gain is $\Delta = (c - b)/n$, with standard error $\text{SE} = \sqrt{(b + c) - (c - b)^2/n}\,/\,n$. A 95\% CI of $\Delta \pm 1.96 \times \text{SE}$ is reported; an asterisk indicates the interval excludes zero.

\begin{table*}[t]
\centering
\small
\begin{tabular}{llrrrrrl}
\toprule
\textbf{Model} & \textbf{Language} & \textbf{Base (\%)} & \textbf{FT (\%)} & \textbf{$\Delta$ (pp)} & \textbf{$b$} & \textbf{$c$} & \textbf{95\% CI} \\
\midrule
\multirow{7}{*}{DeepSeek-V3.1}
  & Amharic  & 75 & 81 & +6  & 0 & 6  & [+1.3, +10.7]* \\
  & Hebrew   & 91 & 94 & +3  & 0 & 3  & [$-$0.3, +6.3] \\
  & Maltese  & 87 & 89 & +2  & 2 & 4  & [$-$2.8, +6.8] \\
  & Japanese & 91 & 93 & +2  & 1 & 3  & [$-$1.9, +5.9] \\
  & Korean   & 92 & 94 & +2  & 0 & 2  & [$-$0.7, +4.7] \\
  & French   & 93 & 92 & $-$1 & 1 & 0 & [$-$3.0, +1.0] \\
  & \textit{Average} & & & \textit{+2.3} & & & \textit{[+0.8, +3.8]*} \\
\midrule
\multirow{7}{*}{GPT-OSS-120B}
  & Amharic  & 30 & 52 & +22 & 12 & 34 & [+9.4, +34.6]* \\
  & Hebrew   & 41 & 85 & +44 & 4  & 48 & [+32.8, +55.2]* \\
  & Maltese  & 32 & 73 & +41 & 7  & 48 & [+28.9, +53.1]* \\
  & Japanese & 55 & 84 & +29 & 5  & 34 & [+18.2, +39.8]* \\
  & Korean   & 56 & 88 & +32 & 0  & 32 & [+22.9, +41.1]* \\
  & French   & 58 & 91 & +33 & 2  & 35 & [+23.0, +43.0]* \\
  & \textit{Average} & & & \textit{+33.5} & & & \textit{[+29.0, +38.0]*} \\
\midrule
\multirow{7}{*}{GPT-OSS-20B}
  & Amharic  & 27 & 50 & +23 & 10 & 33 & [+11.0, +35.0]* \\
  & Hebrew   & 66 & 91 & +25 & 1  & 26 & [+16.1, +33.9]* \\
  & Maltese  & 29 & 73 & +44 & 3  & 47 & [+33.2, +54.8]* \\
  & Japanese & 69 & 89 & +20 & 3  & 23 & [+10.8, +29.2]* \\
  & Korean   & 78 & 91 & +13 & 3  & 16 & [+4.8, +21.2]* \\
  & French   & 82 & 89 & +7  & 2  & 9  & [+0.6, +13.4]* \\
  & \textit{Average} & & & \textit{+22.0} & & & \textit{[+18.1, +25.9]*} \\
\midrule
\multirow{7}{*}{Qwen3-32B}
  & Amharic  & 52 & 55 & +3  & 1 & 4  & [$-$1.3, +7.3] \\
  & Hebrew   & 87 & 88 & +1  & 0 & 1  & [$-$1.0, +3.0] \\
  & Maltese  & 85 & 86 & +1  & 0 & 1  & [$-$1.0, +3.0] \\
  & Japanese & 93 & 94 & +1  & 0 & 1  & [$-$1.0, +3.0] \\
  & Korean   & 90 & 90 & 0   & 0 & 0  & [0.0, 0.0] \\
  & French   & 91 & 91 & 0   & 0 & 0  & [0.0, 0.0] \\
  & \textit{Average} & & & \textit{+1.0} & & & \textit{[+0.1, +1.9]*} \\
\midrule
\multirow{7}{*}{Qwen3-4B-Instruct}
  & Amharic  & 37 & 42 & +5  & 3 & 8  & [$-$1.4, +11.4] \\
  & Hebrew   & 81 & 81 & 0   & 2 & 2  & [$-$3.9, +3.9] \\
  & Maltese  & 54 & 61 & +7  & 7 & 14 & [$-$1.9, +15.9] \\
  & Japanese & 84 & 82 & $-$2 & 3 & 1 & [$-$5.9, +1.9] \\
  & Korean   & 90 & 89 & $-$1 & 1 & 0 & [$-$3.0, +1.0] \\
  & French   & 91 & 90 & $-$1 & 1 & 0 & [$-$3.0, +1.0] \\
  & \textit{Average} & & & \textit{+1.3} & & & \textit{[$-$0.8, +3.4]} \\
\midrule
\multirow{7}{*}{Qwen3-8B}
  & Amharic  & 44 & 44 & 0   & 3 & 3  & [$-$4.8, +4.8] \\
  & Hebrew   & 79 & 79 & 0   & 0 & 0  & [0.0, 0.0] \\
  & Maltese  & 65 & 63 & $-$2 & 3 & 1 & [$-$5.9, +1.9] \\
  & Japanese & 82 & 84 & +2  & 1 & 3  & [$-$1.9, +5.9] \\
  & Korean   & 89 & 89 & 0   & 0 & 0  & [0.0, 0.0] \\
  & French   & 92 & 92 & 0   & 0 & 0  & [0.0, 0.0] \\
  & \textit{Average} & & & \textit{0.0} & & & \textit{[$-$1.2, +1.2]} \\
\midrule
\multirow{7}{*}{Qwen3-8B-Base}
  & Amharic  & 43 & 45 & +2  & 7 & 9  & [$-$5.8, +9.8] \\
  & Hebrew   & 81 & 85 & +4  & 1 & 5  & [$-$0.7, +8.7] \\
  & Maltese  & 72 & 70 & $-$2 & 5 & 3 & [$-$7.5, +3.5] \\
  & Japanese & 83 & 83 & 0   & 3 & 3  & [$-$4.8, +4.8] \\
  & Korean   & 88 & 87 & $-$1 & 3 & 2 & [$-$5.4, +3.4] \\
  & French   & 89 & 91 & +2  & 0 & 2  & [$-$0.7, +4.7] \\
  & \textit{Average} & & & \textit{+0.8} & & & \textit{[$-$1.3, +3.0]} \\
\bottomrule
\end{tabular}
\caption{Paired 95\% confidence intervals for fine-tuning gains using McNemar's test ($n = 100$ per language). $b$ = regressions (base correct, FT wrong); $c$ = improvements (base wrong, FT correct). Asterisks indicate intervals excluding zero. We use this as a descriptive bookkeeping marker rather than a significance test.}
\label{tab:ci_details}
\end{table*}

\section{CoT Truncation Rates}
\label{sec:truncation}

Table~\ref{tab:truncation} reports the percentage of reasoning ablation examples whose CoT generation reached the 4,096-token budget, along with the average and maximum token counts per model--language pair. All truncated examples are retained in the evaluation (their reasoning trace is simply cut off at the token limit). Truncation rates are highest for lower-resource languages (e.g., Amharic) and for smaller models that produce longer reasoning traces.

\begin{table*}[t]
\centering
\small
\begin{tabular}{llrrrr}
\toprule
\textbf{Model} & \textbf{Language} & \textbf{$n$} & \textbf{Truncated (\%)} & \textbf{Avg.\ Tokens} & \textbf{Max Tokens} \\
\midrule
\multirow{7}{*}{DeepSeek-V3.1 (671B)}
  & Amharic  & 100 & 0.0 & 304 & 953 \\
  & Hebrew   & 100 & 0.0 & 164 & 743 \\
  & Maltese  & 100 & 0.0 & 216 & 1,348 \\
  & Japanese & 100 & 0.0 & 223 & 1,692 \\
  & Korean   & 100 & 0.0 & 176 & 1,215 \\
  & French   & 100 & 0.0 & 169 & 488 \\
  & \textit{Overall} & \textit{600} & \textit{0.0} & \textit{209} & \textit{1,692} \\
\midrule
\multirow{7}{*}{GPT-OSS-120B}
  & Amharic  & 100 & 1.0 & 1,247 & 4,096 \\
  & Hebrew   & 100 & 0.0 & 294 & 2,019 \\
  & Maltese  & 100 & 0.0 & 405 & 1,461 \\
  & Japanese & 100 & 1.0 & 473 & 4,096 \\
  & Korean   & 100 & 0.0 & 334 & 1,405 \\
  & French   & 100 & 0.0 & 301 & 976 \\
  & \textit{Overall} & \textit{600} & \textit{0.3} & \textit{509} & \textit{4,096} \\
\midrule
\multirow{7}{*}{GPT-OSS-20B}
  & Amharic  & 100 & 20.0 & 2,116 & 4,096 \\
  & Hebrew   & 100 & 5.0 & 519 & 4,096 \\
  & Maltese  & 100 & 6.0 & 753 & 4,096 \\
  & Japanese & 100 & 5.0 & 662 & 4,096 \\
  & Korean   & 100 & 3.0 & 536 & 4,096 \\
  & French   & 100 & 2.0 & 438 & 4,096 \\
  & \textit{Overall} & \textit{600} & \textit{6.8} & \textit{837} & \textit{4,096} \\
\midrule
\multirow{7}{*}{Qwen3-32B}
  & Amharic  & 100 & 3.0 & 861 & 4,096 \\
  & Hebrew   & 100 & 4.0 & 576 & 4,096 \\
  & Maltese  & 100 & 7.0 & 814 & 4,096 \\
  & Japanese & 100 & 4.0 & 672 & 4,096 \\
  & Korean   & 100 & 2.0 & 599 & 4,096 \\
  & French   & 100 & 6.0 & 680 & 4,096 \\
  & \textit{Overall} & \textit{600} & \textit{4.3} & \textit{700} & \textit{4,096} \\
\midrule
\multirow{7}{*}{Qwen3-8B}
  & Amharic  & 100 & 16.0 & 1,989 & 4,096 \\
  & Hebrew   & 100 & 3.0 & 958 & 4,096 \\
  & Maltese  & 100 & 5.0 & 1,300 & 4,096 \\
  & Japanese & 100 & 6.0 & 1,134 & 4,096 \\
  & Korean   & 100 & 1.0 & 873 & 4,096 \\
  & French   & 100 & 7.0 & 1,017 & 4,096 \\
  & \textit{Overall} & \textit{600} & \textit{6.3} & \textit{1,212} & \textit{4,096} \\
\bottomrule
\end{tabular}
\caption{CoT truncation rates for the reasoning ablation (max budget: 4,096 tokens). Truncated (\%) indicates the fraction of examples whose generation reached the token limit. Qwen3-4B-Instruct and Qwen3-8B-Base are omitted as they do not support CoT inference. Overall truncation across all models is 3.6\% (107/3,000 examples).}
\label{tab:truncation}
\end{table*}

\section{Prompt Template}
\label{sec:prompt_template}

The following prompt template is used for both training and evaluation. The instruction is always in Arabic; the passage, question, and answer choices are in the target language (Arabic dialects during training, and the target evaluation language during zero-shot evaluation). During training, the model's target is the correct answer digit (1--4), with the loss computed only over assistant-turn tokens. During evaluation, the prompt is followed by the anchor string \texttt{"ANSWER:~"} and scored via logprobs as described in Section~\ref{evaluation}.

\begin{verbatim}
[Arabic instruction: "Read the following
 text then answer the question:"]

[Arabic: "Text:"]
{passage}

[Arabic: "Question:"]
{question}

[Arabic: "Choices:"]
1) {choice_1}
2) {choice_2}
3) {choice_3}
4) {choice_4}

[Arabic: "Please choose the correct answer
 number (1, 2, 3, or 4)."]
\end{verbatim}

\noindent The Arabic instruction lines are fixed across all examples and languages. The placeholders \texttt{\{passage\}}, \texttt{\{question\}}, and \texttt{\{choice\_1\}}--\texttt{\{choice\_4\}} are populated from the Belebele dataset in the target language. The entire prompt is wrapped in the model's chat template before tokenization.

\section{Answer Calibration Details}
\label{sec:calibration_details}

This appendix reports the distributional calibration analysis referenced in Sections~\ref{sec:results}, \ref{sec:reasoning_ablation}, and the Discussion. For each model--language pair, we compute three metrics over the four answer-choice probabilities obtained by normalizing the logprobs from Eq.~\ref{eq:logprob}: Shannon entropy $H = -\sum_i p_i \log_2 p_i$ (range $[0, 2]$ bits; lower indicates a sharper distribution), top-2 margin $M = p_{\text{top1}} - p_{\text{top2}}$ (range $[0, 1]$; higher indicates more decisive separation), and $P^*$, the mean probability assigned to the correct answer. Together, these separate distributional sharpening ($\Delta H < 0$, $\Delta M > 0$) from sharpening \emph{toward the correct answer} ($\Delta P^* > 0$).

\paragraph{Summary.}

Table~\ref{tab:calibration_summary} reports average calibration deltas across all six target languages for both fine-tuning and CoT reasoning. The key patterns are: (i) models with high baseline entropy (GPT-OSS-120B, GPT-OSS-20B) show large entropy reductions and $\Delta P^* > 0$ under both interventions, confirming that accuracy gains reflect improved calibration; (ii) models with low baseline entropy (DeepSeek-V3.1, Qwen3-32B) show minimal $\Delta P^*$ despite some distributional sharpening under CoT; and (iii) the model ranking by $\Delta P^*$ is preserved across fine-tuning and CoT, consistent with the hypothesis that both address the same format-alignment bottleneck.

\begin{table}[t]
\centering
\small
\begin{tabular}{lrrrr}
\toprule
\textbf{Model} & $\Delta H$ & $\Delta M$ & $\Delta P^*$ & $\Delta$\textbf{Acc} \\
\midrule
\multicolumn{5}{l}{\textit{Fine-tuning (base $\to$ ft)}} \\
\midrule
GPT-OSS-120B   & $-$0.34 & +0.28 & +0.29 & +33.5 \\
GPT-OSS-20B    & $-$0.63 & +0.24 & +0.23 & +22.0 \\
Qwen3-8B-Base  & $-$0.19 & +0.08 & +0.05 & +0.8 \\
Qwen3-32B      & +0.02  & $-$0.00 & +0.00 & +1.0 \\
DeepSeek-V3.1  & +0.13  & $-$0.02 & +0.00 & +2.3 \\
Qwen3-8B       & +0.12  & $-$0.04 & $-$0.01 & +0.0 \\
Qwen3-4B-Inst. & +0.42  & $-$0.17 & $-$0.01 & +1.3 \\
\midrule
\multicolumn{5}{l}{\textit{CoT reasoning (no CoT $\to$ with CoT)}} \\
\midrule
GPT-OSS-120B   & $-$1.15 & +0.61 & +0.45 & +41.0 \\
GPT-OSS-20B    & $-$0.98 & +0.40 & +0.27 & +22.8 \\
Qwen3-8B       & $-$0.37 & +0.17 & +0.06 & +4.3 \\
Qwen3-32B      & $-$0.33 & +0.13 & +0.02 & +0.3 \\
DeepSeek-V3.1  & $-$0.30 & +0.12 & +0.02 & $-$0.7 \\
\bottomrule
\end{tabular}
\caption{Average calibration deltas across all six target languages. $\Delta H$: change in Shannon entropy (bits; negative $=$ sharper). $\Delta M$: change in top-2 margin (positive $=$ more decisive). $\Delta P^*$: change in mean probability on the correct answer. $\Delta$Acc: accuracy change (pp). Models sorted by $\Delta P^*$ within each panel.}
\label{tab:calibration_summary}
\end{table}

\paragraph{Fine-tuning diffuses already-calibrated models.}

An unexpected pattern among the well-calibrated dense models is that fine-tuning \emph{increases} entropy. Table~\ref{tab:calibration_ft_detail} reports per-language deltas for the most extreme case (Qwen3-4B-Instruct) alongside a representative high-gain model (GPT-OSS-120B). For Qwen3-4B-Instruct, entropy rises on every language, most dramatically on Amharic ($\Delta H = +1.05$) and Maltese ($\Delta H = +0.80$), with $\Delta P^*$ near zero throughout. This indicates that LoRA fine-tuning on Arabic introduces interference that diffuses the model's logprob distributions without redirecting mass toward correct answers---a distributional signature consistent with the near-zero accuracy gains reported in Section~\ref{sec:results}. By contrast, GPT-OSS-120B shows entropy decreases and $\Delta P^* > 0$ across all languages, confirming that its accuracy gains reflect genuine calibration improvement.

\begin{table*}[t]
\centering
\small
\begin{tabular}{llrrrrrr}
\toprule
\textbf{Model} & \textbf{Metric} & \textbf{Amharic} & \textbf{Hebrew} & \textbf{Maltese} & \textbf{Japanese} & \textbf{Korean} & \textbf{French} \\
\midrule
\multirow{3}{*}{GPT-OSS-120B}
  & $\Delta H$  & +0.18 & $-$0.50 & $-$0.30 & $-$0.41 & $-$0.46 & $-$0.57 \\
  & $\Delta M$  & +0.01 & +0.38 & +0.31 & +0.30 & +0.31 & +0.34 \\
  & $\Delta P^*$ & +0.12 & +0.36 & +0.33 & +0.29 & +0.32 & +0.32 \\
\midrule
\multirow{3}{*}{Qwen3-4B-Inst.}
  & $\Delta H$  & +1.05 & +0.21 & +0.80 & +0.21 & +0.13 & +0.11 \\
  & $\Delta M$  & $-$0.43 & $-$0.10 & $-$0.34 & $-$0.08 & $-$0.05 & $-$0.04 \\
  & $\Delta P^*$ & +0.01 & $-$0.01 & +0.02 & $-$0.02 & $-$0.02 & $-$0.01 \\
\bottomrule
\end{tabular}
\caption{Per-language calibration deltas from fine-tuning for GPT-OSS-120B (high gain) and Qwen3-4B-Instruct (near-zero gain). GPT-OSS-120B sharpens toward correct answers on all languages except Amharic, where it starts closest to chance. Qwen3-4B-Instruct shows entropy increases and near-zero $\Delta P^*$ across the board.}
\label{tab:calibration_ft_detail}
\end{table*}

\paragraph{CoT calibration: answer-indicator leakage and residual signal.}

An important caveat applies to interpreting the CoT calibration numbers. Because the CoT protocol generates a free-form reasoning trace before logprob scoring (Eq.~\ref{eq:logprob_cot}), and these traces frequently contain explicit answer indicators (e.g., ``the correct answer is 3''), the model is partly conditioning on its own stated answer when the digit logprobs are computed. This mechanically collapses the distribution toward a single token, explaining the near-zero post-CoT entropies in Table~\ref{tab:calibration_cot_detail} (mean $H = 0.054$ bits for GPT-OSS-120B, $0.001$ for DeepSeek-V3.1). The absolute magnitudes of $\Delta H$ and $\Delta M$ under CoT are therefore not directly comparable to those from fine-tuning, which operates on the same unconditional scoring context as the baseline.

What remains informative is $\Delta P^*$---the change in probability mass on the \emph{correct} answer. Answer-indicator leakage concentrates mass on whichever answer the model stated in its trace, but it only increases $P^*$ if the model reasoned its way to the right answer. Table~\ref{tab:calibration_cot_detail} shows that GPT-OSS-120B achieves $\Delta P^*$ values ranging from +0.33 (Amharic) to +0.60 (Maltese), indicating that its reasoning traces overwhelmingly arrive at correct answers. DeepSeek-V3.1, by contrast, shows $\Delta P^*$ of only +0.02 on average despite equally extreme sharpening---its traces are confident but do not systematically redirect mass toward correct answers that it was not already selecting. This model-level dissociation between $\Delta H$ (large for all models) and $\Delta P^*$ (large only for weak-baseline models) parallels the fine-tuning calibration pattern and is consistent with the format-calibration hypothesis, even after accounting for the leakage confound.

\begin{table*}[t]
\centering
\small
\begin{tabular}{llrrrrrr}
\toprule
\textbf{Model} & \textbf{Metric} & \textbf{Amharic} & \textbf{Hebrew} & \textbf{Maltese} & \textbf{Japanese} & \textbf{Korean} & \textbf{French} \\
\midrule
\multirow{3}{*}{GPT-OSS-120B}
  & $\Delta H$  & $-$1.29 & $-$1.16 & $-$1.32 & $-$1.13 & $-$1.01 & $-$0.99 \\
  & $\Delta M$  & +0.67 & +0.65 & +0.73 & +0.57 & +0.53 & +0.51 \\
  & $\Delta P^*$ & +0.33 & +0.49 & +0.60 & +0.45 & +0.41 & +0.43 \\
\midrule
\multirow{3}{*}{DeepSeek-V3.1}
  & $\Delta H$  & $-$0.76 & $-$0.25 & $-$0.28 & $-$0.21 & $-$0.18 & $-$0.14 \\
  & $\Delta M$  & +0.32 & +0.10 & +0.11 & +0.09 & +0.07 & +0.06 \\
  & $\Delta P^*$ & +0.08 & $-$0.01 & +0.03 & $-$0.01 & +0.03 & +0.02 \\
\bottomrule
\end{tabular}
\caption{Per-language calibration deltas from CoT reasoning for GPT-OSS-120B and DeepSeek-V3.1. Both models sharpen dramatically ($\Delta H \ll 0$), but only GPT-OSS-120B redirects mass toward correct answers ($\Delta P^* \gg 0$). DeepSeek-V3.1's near-zero $\Delta P^*$ despite large $\Delta H$ confirms that strong-baseline models were already correct at the argmax level.}
\label{tab:calibration_cot_detail}
\end{table*}

\end{document}